\documentclass[preprint,12pt]{elsarticle}

\usepackage{amssymb}
\usepackage[inline]{enumitem}

\usepackage{supertabular}
\usepackage{array}
\newcolumntype{L}{>{\centering\arraybackslash}m{3cm}}

\usepackage{listings,lstautogobble}
\usepackage{xcolor}
\definecolor{RoyalBlue}{cmyk}{1, 0.50, 0, 0}

\journal{***}

\begin{document}
%\lstlistoflistings

\begin{frontmatter}



\title{FT-SWRL: A Fuzzy-Temporal Extension of Semantic Web Rule Language}

\author[label1]{Abba Lawan}
\ead{khyx3alw@nottingham.edu.my}
\author[label2]{Abdur Rakib}
\ead{Rakib.Abdur@uwe.ac.uk}
%\author[label2]{Natasha Alechina}
%\ead{Natasha.Alechina@nottingham.ac.uk}

%\address[label2]{School of Computer Science\\ The University of Nottingham, U.K.}
\address[label1]{Crops For the Future Research Centre (CFFRC), Malaysia}
\address[label2]{Department of Computer Science and Creative Technologies\\University of the West of England, Bristol}

%\author{Abba Lawan \corref{cor1}\fnref{label2}}
%\ead{abbalawal2002@gmail.com}
%\fntext[label2]{University of Nottingham Malaysia}
%\cortext[cor1]{Crops for the Future (CFF)}
%\address{Malaysia\fnref{label3}}
%\fntext[label3]{CFF Malaysia}

%\address{}

\begin{abstract}

We present, FT-SWRL, a fuzzy temporal extension to the Semantic Web Rule Language (SWRL), which combines fuzzy theories based on the valid-time temporal model to provide a standard approach for modeling imprecise temporal domain knowledge in OWL ontologies. The proposal introduces a fuzzy temporal model for the semantic web, which is syntactically defined as a fuzzy temporal SWRL ontology (SWRL-FTO) with a new set of fuzzy temporal SWRL built-ins for defining their semantics. The SWRL-FTO hierarchically defines the necessary linguistic terminologies and variables for the fuzzy temporal model. An example model demonstrating the usefulness of the fuzzy temporal SWRL built-ins to model imprecise temporal information is also represented. Fuzzification process of interval-based temporal logic is further discussed as a reasoning paradigm for our FT-SWRL rules, with the aim of achieving a complete OWL-based fuzzy temporal reasoning. Comparative review on fuzzy temporal representation approaches, both within and without the use of ontologies, shows that the FT-SWRL model can thus serve as a formal specification for handling imprecise temporal expressions in domain knowledge modeling.
\end{abstract}

\begin{keyword}
Valid Time Model\sep Fuzzy Temporal Logic \sep Ontology \sep Language Extensions\sep SWRL Built-ins.


\end{keyword}

\end{frontmatter}


\section{Introduction}
\label{intro}

Conceptual domain modeling in the field of the Semantic Web is often achieved through Description Logic (DL)-based ontology languages such as OWL and is typically guided based on the normal set theory. Modeling imprecise temporal expressions (ITEs) that depend on unstructured vague time data is still a challenge and due to their limited syntax and semantics, current domain modeling languages require that temporal information is asserted as definite time-points. However, such limitations may lead to unfounded approximations and quickly translate to a loss of information in modeling the real-world, especially in application ontologies. Hence, there is a need for a comprehensive ontology language for handling temporal uncertainties (vague expressions of time) commonly found in the real-world narration of domain facts.
 
\par As the use of ontologies in enterprise applications is becoming pervasive, effective representation and communicating of fuzzy domain facts cannot be overestimated. As such, various language extensions have been inspired by the fuzzy set theory to enable representing non-crisp or vague facts into ontology models. These extensions were mainly rooted in the fuzzy extension of the underlying description logic, leading to the evolution of both OWL and SWRL language extensions, including the Fuzzy-OWL\cite{Stoilos2005}, Fuzzy-SWRL \cite{Pan2004}, SWRL-Fuzzy \cite{Wlodarczyk2010}, and Vague-SWRL \cite{Wang2008}. Similarly, though with a much lesser magnitude, temporal ontologies and language extensions have been proposed for handling time-related information in the semantic web.  Available ontologies include the OWL Time ontology \cite{Hobbs2004}, which basically describes the general concepts of time as entities. This is due to the logic-based function-free approach of OWL -- meaning that temporal arguments cannot be added to the supported binary relationships. Whereas, available temporal language extensions include the Temporal SWRL among others, which aim to provide basic constructs to describe temporal facts in a knowledge domain with some degree of consistency \cite{Connor2011}. The SWRL-Temporal model provides a standard mechanism for representing and managing temporal information based on the Valid-time temporal model – commonly used to represent temporal information in knowledge-based systems \cite{Snodgrass1996adding}. 

However, there still exists a wide research gap in achieving consistent representation formalisms for managing temporal uncertainties in domain ontologies. Modeling imprecise temporal expressions that depend on unstructured vague time data is still a challenge in the field of semantic web. Moreover, considering the diverse nature of information on the web (and by extension, the semantic web) --- which to some considerable proportion involves experts as well as novice's narratives of events, the advancement of the semantic web no doubt requires an even more expressive modeling formalisms.  
The purpose of this research, therefore, is to bridge this gap by introducing a consistent fuzzy-temporal extension that can be used to represent and reason over uncertain-temporal domain knowledge in the semantic web. The main objective of the paper is to propose an extension of SWRL, the standard semantic web rule language in order to deal with imprecise temporal expressions. The paper proposes to represent imprecise temporal expressions as fuzzy intervals.

As described in \cite{Ziqiang227942}, there are two types of temporal uncertainties --- the imprecise dating of events and the fuzzy description of temporal data. The FT-SWRL model thus aims to provide a new formalism for representing the latter in ontologies (i.e. modeling fuzzy temporal data) using SWRL rules. The new extension is defined as a new fragment of the existing SWRL Temporal formalism, where a fuzzy temporal ontology has been developed to extend the temporal ontology and new set of built-ins defined to represent the semantics of the imprecise temporal expressions for the reasoning purposes.  
\par 
By defining the fuzzy temporal SWRL ontology (SWRL-FTO) as a reference model and designing relevant built-in operators, the FT-SWRL extension will surely improve the usability of the existing SWRL temporal formalism. While the basic temporal SWRL rules can represent interval operators such as those described as Allen’s temporal operators \cite{Allen1983}, utilizing such operators will be incomplete without some degree of fuzziness in the knowledge base. Moreover, fuzzification of imprecise temporal expressions from a single time stamp or interval to a more realistic set of possible time intervals, usually results in a wider range of temporal operations, such as \textit{overlaps, meets}, and  \textit{contains}.  Regardless of the overheads, such modeling scenario confirms the assertion that FT-SWRL does not only help to represent fuzzy temporal information in OWL ontologies but can also help to improve the utilization of its existing temporal model operators. 

\subsection{Motivation}
\label{sec:motivation}
As stated earlier, the aim of this research is to provide a consistent representation model for modeling both the temporal data and the temporal uncertainties commonly found in domain facts. This is achieved by extending the formalisms of the Semantic Web Rule Language with fuzzy temporal constructs, syntactically defined as in a fuzzy temporal SWRL ontology model and built-ins. As such, FT-SWRL enabled ontologies are expected to consistently model such dynamic and uncertain phenomena that are otherwise difficult to manage using the basic OWL/SWRL constructs or temporal models. For instance, in the current OWL/SWRL definitions, expressions involving approximate temporal assertions such as \textit{around 4pm, about 4 hours, few weeks ago,} etc. cannot be easily represented nor efficiently extracted from OWL ontologies. 
\par 
In essence, the FT-SWRL extension is aimed at providing such syntactical and semantic extensions in the existing SWRL formalism for handling imprecise temporal information by providing fuzzy-grounded classes and property constructs, built-ins, and annotation properties to encode fuzzy-temporal information in the SWRL formalism.
We extend the SWRL-Temporal model with such commonly utilized Imprecise Temporal Expressions (ITEs) found in descriptions of domain facts. This is particularly important where the knowledge to be represented is in the form of expert opinions or natural language narratives of experts, with no prior knowledge of ontological domain modeling. As in the motivational case study \cite{Lawan2014}, where we use SWRL-enabled ontologies to model the farming practices of underutilized crops --- a field that significantly relies on local farmers’ expertise alongside the scientific knowledge. Consider the representation of following facts on underutilized crops (Bambaranut) domain:

\begin{itemize}\label{fragmentofBambaraFacts}
	
	\item	Bambara groundnut requires a growth period of \textit{about 110 to 150 days} for the crop to be developed.
	\item	Bambara beans take \textit{around 7 to 15 days} to germinate. Seed stored for \textit{about 12 months} germinate well, but longer storage results in loss of viability. 
	\item	Flowering \textit{starts 30 to 35 days} after sowing and may continue \textit{until} the end of the plant’s life.
	\item	Pod and seed development take place \textit{approximately 30 to 40 days} after fertilization. This takes \textit{up to 30 days after} fertilization. The seed develops \textit{during a further 10 days}. \cite{Swanevelder1998}
	
\end{itemize}

While handling time-related data alone or managing uncertainties in a domain knowledge are in themselves difficult tasks. Nonetheless, there is a need to model such real-world issues that require a representation of time changes and the uncertainties brought about by these changes or temporal relations between events. For example, a given class (GrowthStage) in our Underutilized-Crops ontology\cite{Lawan2014} can have different individual instances asserted depending on the planting date of a crop (represented as \textit{hasDateOfSowing} datatype property). However, temporal measurements and time itself are not fixed or definite data and therefore the time inputs are merely estimates. As such the use of descriptors, such as \textit{about}, \textit{approximately}, \textit{around} followed by a time value merely confirms that such information is imprecise --- and hence the need to be modeled as such. Other common application areas of fuzzy-temporal modeling for handling imprecise temporal expressions include the medical domain, multimedia, market trends analysis, and natural language applications in virtual assistants (e.g., Siri in Apple, Cortana by Microsoft and Google-Now), among others.

\subsection{Scope}
Since it is such an enormous task to generalize all fuzzy set theories into an ontology rule language extension, we adopt the bottom-up approach where we begin with introducing fuzziness from the peripherals of SWRL by extending existing temporal model. This has the advantage of working with existing tools during implementation and without introducing inconsistencies to main ontologies. As such, our proposal does not focus on modeling rudimentary time concepts and terminologies but relies mainly on existing standards that are already compatible with SWRL, such as the OWL time ontology and XML schema temporal data types --- the xsd:(date, dateTime, and duration), among others. Moreover, the SWRL temporal model, presented in \cite{Connor2011}, serves as the basis for the adopted crisp valid-temporal model. In essence, our primary focus is providing an extended abstract syntax and semantics for fuzzy temporal representation in the SWRL language. This allows modeling imprecise temporal facts using fuzzy temporal modifiers (a collection of fuzzy terms and variables) defined as constructs. We focused on the Semantic Web Rule Language (SWRL) and its fuzzy temporal extension largely due to its semantic integration with OWL and the ability of SWRL to assert domain knowledge into ontologies, as well as extract them using its query functionality – available as an SQWRL (SWRL query language) \cite{OConnor2009}. 
\par
In what follows, we presented the preliminaries in Section~\ref{prelim} to serve as introductory notes for those new to SWRL, fuzzy logics, and temporal models. In Section~\ref{FTSWRL}, we describe a complete fuzzification process by first defining the SWRL-FTO ontology and a new set of SWRL built-ins to formally specify the linguistic terminologies and variables of the FT-SWRL model. This is followed by a discussion of the reasoning paradigm for the new formalism in Section~\ref{reasoningparadigm}. For completeness, we discuss the relevant works in fuzzy temporal representation and reasoning in Section~\ref{relevantworks} and conclude in Section~\ref{conclusion}.

\section{Preliminaries}
\label{prelim}
\subsection{Valid-Time Temporal Model}
A valid time temporal model \cite{Snodgrass1996adding} helps to provide a simple and consistent approach for modeling temporal information (called facts or propositions). In this model, a temporal proposition is either true or valid as specified in its associated timestamp --- called the valid time. Such timestamps can be either specific time instants or intervals (time period between time instants). A special time-interval called Duration is characterized by two arguments: the Granularity --- which is a unit measure for temporal data e.g,. days, hours, seconds, etc. and the Duration count --- usually an integer. However, Duration can also be specified using two 'time instants' (with xsd:dateTime as arguments). \par

From the literature, basic temporal objects have been commonly classified into three distinctive references as follows: \begin{enumerate*}[label=(\roman*)]
	\item Points in time – which defines a single temporal point on the timeline. Example: 13:00, now, date, etc.
	\item Time Intervals – defining the temporal relationship between two time-points. Example: 13:00 – 14:00, 2–3 December, etc.
	\item Duration or relative expression of intervals. Example: 2 weeks, 6 years, many hours, etc. usually represented as counts of a time granularity.
\end{enumerate*}  
 \par Based on the valid-time temporal model, the Fuzzy Temporal SWRL model was proposed in \cite{Connor2011} by defining two important entities: the SWRL Temporal Ontology and the SWRL Temporal Built-ins. We briefly discuss them below.
 
\subsection{SWRL Temporal Ontology}
\begin{figure}[th]
	\centering
	\includegraphics[width=0.55\linewidth, height=0.30\textheight]{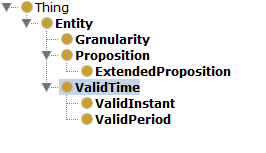}
	\vspace{-0.5cm}
	\caption[SWRL Temporal Ontology showing ValidTime class hierarchy]{SWRL Temporal Ontology showing ValidTime class hierarchy}
	\label{fig:swrltemporalontologyfrag}
\end{figure}

The SWRL temporal ontology\footnote{http://swrl.stanford.edu/ontologies/built-ins/3.3/temporal.owl} defines the OWL constructs that can be used to represent the valid-time temporal model. It hierarchically defined the collection of entities that allow modeling interval-based temporal information in OWL ontologies. It has the default prefix: temporal and in its complete form, the ontology also defines the built-ins (SWRL temporal built-ins) for processing and reasoning about the SWRL valid-time temporal model. In the ontology (see fragment view in Fig. \ref{fig:swrltemporalontologyfrag}), a temporal fact is represented as an 'Extended Proposition' to separate the temporal ontology from the domain ontology. This helps to maintain consistency and allows easy manipulation of the temporal fragment of the ontology without affecting the main ontology. 

The ontology further defines individuals for the granularity class to include (Years, months, days, hours, minutes, seconds and milliseconds) and a set of built-ins that can be used in the SWRL rules to perform temporal reasoning in OWL ontologies. Built-ins defined in the ontology can be classified into three categories: \begin{enumerate*}[label=(\roman*)]
	\item The duration operators --- for reasoning about time durations, with constructs: \textit{duration, durationLessThan, durationEqualTo, durationGreaterThan} and their inverses), 
	\item the standard Allen temporal operators --- for reasoning about qualitative temporal information using the calculus of binary relations on intervals. The built-in implementation has constructs, such as \textit{equals, before, after, meets, metBy, overlaps, overlappedBy, contains, during, starts, startedBy, finishes, finishedBy,} and their inverses, and  
	\item the add operator --- implemented to achieve addition and subtraction comparison of time intervals. The construct is written as \textit{temporal:add}.
\end{enumerate*}
It should be noted here that, in the standard interval calculus, all intervals are assumed to have proper and distinct time, with clear-cut beginnings and ends. This limitation forms the basis of our study, as temporal events and time-related processes in the real-world are usually non-crisp and inexact. And representing them as exact facts will no doubt result in logically inefficient models and knowledge-based systems relying on such information can never be sufficiently intelligent.

\subsection{SWRL Temporal Built-ins}
\label{swrltemporalbuiltins}

A powerful feature of the SWRL formalism, is the ability to extend its definition with user-defined methods for writing application-specific rules. Similar to functions used in rule engines, the SWRL built-ins are predicates that accepts one or more arguments and operate on them during rule execution. Due to the limited temporal support in both OWL and SWRL, the SWRL Temporal Built-in Library is added as an expressiveness extension to the original SWRL definition. Defined as part of the SWRL-API's built-in libraries, the temporal built-ins are hierarchically defined as part of the SWRL temporal ontology. The temporal built-ins provide a rich set of temporal operators designed to allow temporal operations on information described using the temporal ontology. Thus, the built-ins allow temporal reasoning about OWL ontologies using SWRL rules. 

\emph{Syntax and Semantics:} In the basic mode, SWRL temporal built-ins operates on arguments supplied by the XML Schema’s 'date' and 'dateTime' data types – provided as xsd:String with values, such as second, hour, day, time, week, month, and year. These were also defined in the basic OWL temporal ontology (OWL Time)\footnote{https://www.w3.org/TR/2016/WD-owl-time-20160712/}. Whereas, in the advanced mode, the SWRL temporal built-ins work on time information that is completely encoded using the valid-time temporal model. As an example, a rule that asserts a 'Fellow' membership rank to existing work-group members, with registration dates before the year 2000, can be written as: \newline \newline
{Workgroupmember(?m), hasRegDate(?m, ?rgd), temporal:before(?rgd, '2000') $\longrightarrow$  FellowMembers(?m)}.
\label{eq1}

\subsection{Fuzzy Sets and Membership Functions}
\label{fuzzysets}

In contrast to probability theory, the Fuzzy theory is a generalization that studies and facilitates analysis of uncertainties in systems where such uncertainties are born due to vagueness (fuzziness) in the available domain knowledge --- rather than due to randomness (probability) alone \cite{KARWOWSKI1986, Zadeh1965}. The logic being that assertion of 'truth' or otherwise of a given fact can be represented by a varying degree on the closed interval [0, 1] --- denoting the classical false and true values. A pool of real numbers denoted by (0, 1) in-between the interval represents the varying degrees of truth (w). Consequently, Fuzzy Sets \cite{Zadeh1965} have been widely used for modeling uncertainties, where knowledge of a domain is incomplete or marred with vagueness. In contrast to crisp set theory, where an object simply belonged to a given set or otherwise, in fuzzy set theory, membership to a set is subjected to the given weight or degrees of truth (w). In fuzzy conceptualization, objects can belong (or otherwise) to a given class with some degree of certainty. We briefly highlight the formal definitions of fuzzy sets as follows: \newline

\emph{Definition 1:}
In a classical set theory, the membership function ($\mu$ or MF) of an element (x) belonging to a given set (A) is represented thus:  
\begin{equation}\label{key}
\mu A(x)= 1 \iff X \in A, 0  \iff X \notin A.
\end{equation}

However, in a fuzzy set theory, there is more to this crisp representation, where an additional information is provided to denote the degree of certainty that the element (x) belonged or otherwise to the given set (A). In such cases, the fuzzy membership function is written as:
\begin{equation}
\mu A(x)= 1 \iff X \in A,\ 0  \iff X \notin A,\ w\ if\ X\ partially\ belongs\ to\ A.
\end{equation}

Where $w$ is a weighted degree function such that $0 < w < 1$. 
\par
In essence, the membership function $\mu$A(x) is continuous in a fuzzy set theory with a range of [0, 1] = w called the \textit{degree of truth for the membership}. An arbitrary curve is usually designed to represent the input space, also called \textit{the universe of discourse} and the mapping of the membership value on this input space is the membership function ($\mu$). The following also hold true in a fuzzy set theory:
\begin{equation}
\mu_A^{-}(x)= 1 - \mu_A^{-}(x).\end{equation}
\begin{equation}
\mu_{A\cap B}(x) = \min (\mu_A(x), \mu_B(x)).\end{equation}
\begin{equation}
\mu_{A\cup B}(x) = \max (\mu_A(x), \mu_B(x)).\end{equation}

We define a Fuzzy Temporal extension of SWRL as a fusion of these basic concepts with the aim of introducing a fuzzy representation and reasoning model for temporal information using the SWRL rules. Our work focused on the complete stages of the fuzzification process, as defined in \cite{Miliauskaite2014}, where we first define the linguistic terminologies and variables in the SWRL-FT ontology and new set of fuzzy temporal Built-ins – presented in Section~\ref{FTSWRL}. We chose suitable membership functions for some selected imprecise temporal expressions and demonstrate how we can generate their corresponding fuzzy values based on the membership functions as presented in Section~\ref{reasoningparadigm}.

\section{The Fuzzy Temporal Extension of SWRL (FT-SWRL Model)}
\label{FTSWRL}

While fuzzy temporal knowledge modeling has been around since the early days of AI, the semantic web domain has seen fewer advancements in the temporal uncertainty modeling. Relevant research efforts have focused mostly on the uncertainty management or the representation of the temporal data as a domain. In FT-SWRL extension, we go beyond the simple structured time data in ontologies to provide additional syntax and semantics that enable the representation of vague temporal facts in the semantic web. The new semantic web rule language extension can handle the modeling of uncertainties that exist in the time domain, defined in the SWRL-FTO ontology and with the possibility of reasoning and inference through the SWRL fuzzy temporal built-ins. Both presented in this section and followed with example FT-SWRL rules.

\subsection{The SWRL Fuzzy Temporal Ontology}
\label{SWRLFTontology}

The fuzzy-temporal extension of the SWRL language is designed to support modeling and reasoning with imprecise temporal expressions (ITEs) in OWL ontologies. To this end, a fuzzy temporal ontology has been developed to define a consistent model that can be used to represent all fuzzy-temporal facts. The ontology also includes the definition of relevant SWRL built-ins, which extends the existing SWRL built-in library, to allow reasoning operations about the modeled fuzzy-temporal information. As explained earlier, the original SWRL temporal model follows the valid-time temporal model --- where temporal facts are modeled as intervals of time-points. Hence, as the name implies, the FT-SWRL extension basically extends the syntax and semantics of the SWRL temporal model (more specifically the 'Advanced SWRL Temporal model') with appropriate fuzzy syntax and semantics.
\par 
While classical SWRL temporal ontology serves as a reference standard for modeling crisp temporal information, the SWRL-FTO ontology is its carefully extended version with the ability to handle imprecise temporal expressions in domain knowledge representation. The ontology has a default prefix: \textit{fuzzytemporal}, and the hierarchical representation of the ontology is presented in Listing~\ref{listing1}  below:

\begin{lstlisting}[caption=The SWRL Fuzzy-Temporal Ontology (SWRL-FTO), label=listing1]

owl:Thing
owl:Entity
owl:Proposition
	temporal:ExtendedProposition \equiv TemporalProposition	(Time related event class)
		temporal:hasValidTime		//object property
		temporal:hasDuration		//object property
	temporal:ValidTime	(Valid Crisp times of events: instants or period)
			temporal:hasDuration	//object property
			temporal:hasGranularity	//object property
		temporal:ValidInstants	(event occurs at a single instant)

			temporal:hasTime(xml:dateTime)	//Datatype property
		temporal:ValidPeriod (event occurs over an interval of time)
			temporal:hasStart(xml:dateTime)	//Datatype property
			temporal:hasFinish(xml:dateTime) //Datatype property
	temporal:Duration	(Temporal Expressions denoting interval-based temporal information)
		temporal:hasCount (xml:Integer)
		temporal:hasGranularity			
	temporal:Granularity	(years, months, days, hours, minutes, secs, milliseconds)
								(Temporal)
___________________________________________________________________________
						    	   (Fuzzy temporal)
						    	   
	fuzzytemporal:FuzzyTemporalProposition	(Vague temporal fact)
		fuzzytemporal:hasFuzzyTime	//object property
		fuzzytemporal:hasFuzzyModifier	//object property
		fuzzytemporal:hasFuzzyDuration	//object property
	fuzzytemporal:FuzzyTime	(Vaguely known time data)
			fuzzytemporal:hasFuzzyDuration	//object property
		fuzzytemporal:FuzzyTimeInstant \equiv FuzzyTimePoint
		fuzzytemporal:FuzzyTimePeriod 
			fuzzytemporal:minFuzzyTime
			fuzzytemporal:maxFuzzyTime
	fuzzytemporal:FuzzyDuration	(Vague interval-based temporal information)
		fuzzytemporal:hasFuzzyCount	//object property
		fuzzytemporal:hasFuzzyGranularity	//object property
	fuzzytemporal:FuzzyCount	(cycles, times, twice, several, many, long-time, this, next, last, etc.) 
		temporal:hasCount (XML:Integer)	//Datatype property
		temporal:hasGranularity (temporal:Granularity)	//Datatype property
	fuzzytemporal:FuzzyGranularity	(weeks, weekend, fortnight, quarter, noon, etc.) 
		fuzzytemporal:SetGranularity	(Yearly, Monthly, Weekly, daily, hourly, perMinute, 
		perSeconds,  perHour, perWeek, perYear)
		fuzzytemporal:DateGranularity	(past, present, currently, recently, nowadays, ago, since, lately, earlier, etc.)
	fuzzytemporal:FuzzyModifiers 	(Imprecise Temporal Expressions - ITEs e.g. about, around, approx, within, a few, several, many, until, always, very).
		fuzzytemporal:hasWeightedValue \equiv hasWeightDegree)[0,1]	// Functional datatype prop.
		fuzzytemporal:hasMembershipFunction (args: weighted sum)	//datatype prop.
	fuzzytemporal:WeightValues		(0 < w < 1) // Possible Weight Intervals
	fuzzytemporal:MembershipFunction
		fuzzytemporal:mfName	(gaussmf, sigmoidmf, gbellmf, etc.) 
		fuzzytemporal:mfCurve	(plots of membership functions)	
		

\end{lstlisting}

The Listing \ref{listing1} above defines the OWL entities as a reference model for representing fuzzy temporal domain knowledge in OWL ontologies. It also shows the hierarchical layout of the OWL entities with the type of relationships that exists between them. For obvious reasons, it began with the original temporal entities defined in the SWRL temporal model followed by the extended fuzzy-temporal ones defined as the SWRL fuzzy-temporal model. These include the fuzzy temporal proposition, the fuzzy modifiers and their membership function, fuzzy granularity, fuzzy counts, fuzzy time instants and fuzzy durations. In what follows, we briefly highlight these entities and the intuitions behind them.
\paragraph{Representing Fuzzy-temporal Facts:} In order to preserve the modular feature of the original temporal model, the SWRL Fuzzy-temporal ontology (SWRL-FTO) begins with the \textit{FuzzyTemporalProposition} class, which is the class for all fuzzy-timed events i.e., events associated with imprecise temporal expressions. This is defined as a sibling of the \textit{temporal:ExtendedProposition} class --- designed to represent entities or propositions that extend over time and with the benefit of separating the temporal ontology from the main ontology for consistency. Similarly, the \textit{FuzzyTemporalProposition} class will allow introducing consistent fuzzy model without interfering with either the main or temporal ontology. Hence it serves as the range of all the fuzzy temporal built-in expressions defined in the ontology. The \textit{FuzzyTemporalPropositionclass} has three object properties: \textit{hasFuzzyTime} --- with a range over the FuzzyTime class, the \textit{hasFuzzyDuration} --- with a range over the FuzzyDuration class and the \textit{hasModifier} property --- with a range over the FuzzyModifier class.
\newline

The \textit{FuzzyTime} class represents the fuzzy time values of the fuzzy temporal propositions based on the evaluated fuzzy modifiers or ITEs. A FuzzyTime can be either a \textit{FuzzyTimeInstant} or a \textit{FuzzyTimePeriod} --- with a range over the \textit{xsd:dateTime} class. The two describe instantaneous as well as period-based events. Where the time of occurrence of an instantaneous or single timed event is imprecise, then we use the FuzzyTimeInstant (sameAs: FuzzyTimePoint) and FuzzyTimePeriod is used where the event happens over two imprecisely-timed points (period).  \textit{FuzzyDuration} class helps to represent such ITEs containing durations not defined as dateTime data, e.g. 'this weekend', 'within 3 weeks', 'after several hours' etc. As such, it has two subclasses as \textit{FuzzyCounts} and \textit{FuzzyGranularity}. Those granularities that collate their corresponding:

\textit{FuzzyTemporalPropositions} as sets of temporal facts are categorized as \textit{SetGranularities} (e.g. weekly, perHour, daily, etc.) --- a subclass of the FuzzyGranularity class. Whereas, those that basically compare the object propositions with a current date (e.g. past 3 weeks, since last year, 2 weeks ago, etc.) are sub-classed as \textit{DateGranularities}. 
   
\textit{TheFuzzyModifier} class represents the fuzziness of specific fuzzy temporal information or ITEs contained in the fuzzy temporal propositions. It has two object properties: \textit{hasWeightedValue} and \textit{hasMembershipFunction} which range over the \textit{WeightValues} class and \textit{MembershipFunction} class respectively. Corresponding values of the membership function and weight values are assigned to each ITE as functional datatype properties. It represents fuzzy functions for such ITEs as \textit{about, around, approx, within, a few, several, many, until, always, very, etc.} --- and are defined as a set of SWRL built-ins that can be used for temporal operations on the entities defined by the fuzzy temporal ontology. We briefly discuss these built-ins in the preceding section.
A summary of the fuzzy temporal entities is presented in Table \ref{tab:summaryofSWRLentities}: Fuzzy Temporal Classification with their properties, and in Table \ref{tab:SWRLentitiesDR}: Fuzzy Temporal Relations – highlighting their types, domain, and range.

\begin{table}[htbp]
	\centering
\begin{tabular}{p{4cm}|p{4cm}|p{4cm}}
%{|c|c|c|}
	\hline 
	Fuzzy Temporal Class & Sub-classes & Properties \\ 
	\hline \hline 
	
	FuzzyTemporal-Proposition &  & hasFuzzyTime\newline
	hasFuzzyModifier\newline
	hasFuzzyDuration\newline \\ 
	FuzzyTime &	FuzzyTimeInstant
	FuzzyTimePeriod \newline& hasFuzzyDuration\\
	FuzzyTimePeriod	& minFuzzyTime\newline
	maxFuzzyTime\newline&  \\
	FuzzyDuration&  & hasFuzzyCount
	hasFuzzyGranularity\newline\\
	FuzzyCount	& 	& 	hasCount
	hasGranularity\newline\\
	FuzzyGranularity & SetGranularity
	DateGranularity	\newline &  \\
	FuzzyModifiers 	&  & hasWeightedValue
	hasMembershipFunction\newline\\
	WeightValues&  &  \\		
	MembershipFunction  &	mfName \newline
	mfCurve	&   \\
	\hline 
	
\end{tabular} 
\caption{Summary of SWRL Fuzzy Temporal Entities}
\label{tab:summaryofSWRLentities}
\end{table}

Note that, even though the essence of the FT-SWRL extension is to handle imprecise temporal extensions found in domain language narratives, we still introduce some added concepts (e.g. week, quarter, and fortnight) to the original SWRL temporal ontology. Moreover, to accommodate our new constructs, new container classes need to be added leading to the design of a new fuzzy-temporal ontology from scratch.  Hence the above description is that of the fuzzy temporal ontology containing an extended temporal ontology that is set towards modeling natural language description of domain knowledge in OWL ontologies. This approach, will no doubt allow flexible modeling of time-related events. 

\begin{table}[htbp]
	\centering
	\begin{tabular}{p{3.1cm}|p{3cm}|p{3cm}|p{3cm}}
		%{|c|c|c|}
		\hline 
		Fuzzy Temporal Relation & Role Type & Domain & Range \\ 
		\hline \hline 
		
		hasFuzzyTime &	Object property	& FuzzyTemporal-Proposition\newline &	FuzzyTime
				 \\ 
		
		hasFuzzyModifier &	Object property &	FuzzyTemporal-Proposition\newline &	FuzzyModifiers\\
		
		hasFuzzyDuration &	Object property &	FuzzyTemporal-Proposition
		FuzzyTime \newline&	FuzzyDuration \\
		
		hasFuzzyCount &	Object property &	FuzzyDuration	FuzzyCount \newline& \\
		
		hasFuzzyGranu-larity \newline&	Object property &	FuzzyDuration &	FuzzyGranularity \newline\\
		
		hasWeightedValue &	Datatype prop.
		(Functional) \newline&	FuzzyModifiers &	WeightValue	(xml:Decimal) \\
		
		hasMembership-Function\newline &	Datatype prop. &	FuzzyModifiers &	MembershipFunction  \\
		
		hasCount&	Datatype prop. &	FuzzyCount &	xml:Integer \newline\\
		
		hasGranularity &	Datatype prop. &	FuzzyCount	& temporal:Granularity		\\
		\hline 
		
	\end{tabular} 
	\caption{SWRL Fuzzy Temporal Relations summary showing Domain and Range}
	\label{tab:SWRLentitiesDR}
\end{table}

\subsection{SWRL-FT Built-ins: Semantics Definition}
The SWRL fuzzy temporal built-ins are defined to allow temporal operations on imprecise temporal information during domain knowledge modeling. By defining selected ITEs as part of the SWRL built-in sets, FT-SWRL extends SWRL formalism, and equally the OWL language, with constructs to implement fuzzy temporal semantics within ontologies. This will allow the combination OWL/SWRL to handle fuzzy temporal knowledge for the first-time, without relying on external frameworks for reasoning over imprecise temporal data. Using the  FT-SWRL built-ins, imprecise temporal data can be encoded following the SWRL-FTO ontology model and processed based on the valid-time temporal model for efficient knowledge representation and retrieval. Following the fuzzy temporal entities classification in Section \ref{SWRLFTontology}, we define the following keywords as the first set of the SWRL fuzzy-temporal built-in library. Nevertheless, these can be further expanded with more predicates as far as the tractability and semantics of the language can allow.

\subsubsection{Fuzzy Duration Built-ins}
The Fuzzy Duration built-ins were defined to operate on imprecise temporal durations. In this context, \textit{FuzzyDuration} is considered as a temporal expression containing fuzzy Count at a specified base granularity. Unlike the \textit{fuzzytemporal:FuzzyTimePeriod} which can be specified by two fuzzy times instants (\textit{fuzzytemporal:FuzzyTimeInstants}), the fuzzy duration involves expressions, including `few weeks' and `several hours' where the first part (few, several) are the fuzzy counts and the latter (weeks, hours) are the base granularity of the receiving proposition. As such, the FuzzyDuration built-in method requires a \textit{FuzzyCount} and \textit{FuzzyGranularity} as its arguments.
\par
Other operators associated with the FuzzyDuration include the
\emph{fuzzyDurationLessThan, fuzzyDurationEqualsTo}, and \emph{fuzzyDurationGreaterThan} built-ins: As a sub-built-in of the \textit{temporal:Duration} predicate, the FuzzyDuration built-ins inherently includes these operators for comparable inference among consistent FuzzyDuration instants having bounded arguments. Moreover, inverses of these built-ins may well be considered for completeness.

\subsubsection{Fuzzy Count Built-ins}
These built-ins are designed to implement the imprecise counts on temporal data. Example cases include `several', `many', `long-time', `this', `next', `last', `cycles', `times', and `twice'. In their basic form, usage of these built-ins requires that they take the \textit{FuzzyGranularity} as argument and after applying the relevant fuzzy operations defined by their semantics, returns a multiplier or comparison count of the granularity.

\subsubsection{Fuzzy Granularity Built-ins} 
These built-ins are designed to implement the imprecise granularities for an xsd:dateTime class. Example cases include `weeks', `weekend', `fortnight', `quarter', and `noon'. They extend the original SWRL Date, Time and Duration built-ins.\footnote{http://www.daml.org/2004/04/swrl/builtins.html\#8.5}

\subsubsection{Fuzzy Set Granularity Built-ins} 
These built-ins are designed to implement the imprecise set granularities for the xsd:dateTime class. They extend the \textit{FuzzyGranularity} Built-ins to denote set-wise granularities for recurring events. Example cases include, `Yearly', `Monthly', `Weekly', `daily', `hourly', `perMinute', `perSeconds', `perHour', `perWeek', and `perYear''.

\subsubsection{Fuzzy Date Granularity Built-ins} 
These built-ins are designed to implement the imprecise date granularities for annotating the xsd:dateTime class. They extend the \textit{FuzzyGranularity} Built-ins for comparison operations between the current time and the transaction time (temporal proposition object). Example cases include, `the past', `present', `currently', `recently', `nowadays', `ago', `since', `lately', and `earlier''. Hence it requires two optional arguments; the event time and the current time --- in xsd:dateTime instant or duration.

\subsubsection{Imprecise Temporal Approximation Built-ins}  As their name implies, these built-ins are designed to implement the vague temporal approximations. Example cases include, `about', `around', `approx', `within', `a few', `until', `before', `very', and `after'. The built-ins take argument representing the \textit{FuzzyTime} of the temporal fact to apply the relevant fuzzy temporal operations on them. The operation also requires two more arguments representing the \textit{Count} and \textit{FuzzyGranularity} as follows:
\textit{fuzzytemporal:about}(?FuzzyTime, ?Count, ?fuzzyGranularity).
	
\par However, where the Count or Granularity arguments are missing, the default count value is $1$ and a base granularity of the temporal fact will be used. Similarly, inverses of these built-ins (where applicable) can be considered as future extensions.

\subsection{FT-SWRL Rules Usability and Examples}
\label{FTSWRLexamples}

As described in \cite{Connor2006mechanism}, user-defined SWRL built-ins can be used directly in SWRL rules. However, in order to use them and/or their extensions, such as the SWRL-FT built-ins, they need to first be imported into the main ontology by importing their definition --- in this case, the SWRL-FT ontology. Final implementation of the SWRL-FTO ontology requires defining the fuzzy temporal built-ins as instances of the original \textit{swrl:Builtin} class. This is followed by their corresponding Java implementations through the \textit{SWRLBuiltInBbridge} \footnote{http://protege.cim3.net/cgi-bin/wiki.pl?SWRLBuiltInBridge\#nid88T}. The SWRLBuiltInBridge is a component of the SWRLTab (available in the Prot{\'e}g{\'e} ontology editor) that allows the manipulation of SWRL built-ins using Java. Relevant built-ins are usually grouped together in a single OWL file --- which can then be imported into any domain ontology for utilization. 
\par Detailed discussion on the fuzzification of the fuzzy temporal built-ins is presented in Section \ref{reasoningparadigm}. However, in what follows we give some example FT-SWRL rules by modeling some facts about the Bambara groundnut crop --- to highlight possible usability of the built-ins library.

\begin{enumerate}
	\item "Bambara beans take around 1 to 2 weeks to germinate" \newline
	\textit{BambaraBeans (?bb), hasGerminationTime(?bb, ?gt), fuzzytemporal:} \textit{ \textbf{around}(?gt, '2', temporal:weeks) $\longrightarrow$ GermintionPeriod(?bb, True)}.
	
	\item "Bambara beans germinate within 15 days from the date of sowing" \newline
	\textit{BambaraBeans (?bb), hasDateOfSowing(?bb, ?dos), fuzzytemporal:\textbf{within}}  \textit{(?dos, '15', temporal:days) $\longrightarrow$ GerminationPeriod(?bb, True)}
	
	\item "Bambara beans will germinate in few weeks if moderate rainfall continues" \newline
	\textit{BambaraBeans (?bb), hasModerateRainfall(?bb, ?mdR), hasGerminationTime(?bb, ?gt), fuzzytemporal:\textbf{few}(?gt, temporal:weeks) $\longrightarrow$ GerminationPeriod(?bb, True)}
	
	\item "Seeds stored for about12 months germinate well, but longer storage results in loss of viability"\newline
	\textit{Seed (?s), hasStorageTime(?s, ?st), fuzzytemporal:\textbf{before}(?st, 12, temporal:months) $\longrightarrow$ GerminationPeriod(?bb, True)}.\newline
	
	\textit{Seed (?s), hasStorageTime(?s, ?st), fuzzytemporal:\textbf{after}(?st, 12, temporal:months) $\longrightarrow$ GerminationPeriod(?bb, False)}
\end{enumerate}

\section{Reasoning Paradigm for FT-SWRL Rules}
\label{reasoningparadigm}

The original SWRL temporal extension basically defines temporal interval operations as built-ins and neither contain inference rules for time expressions nor translation rules from natural language expression to times.  However, the FT-SWRL proposal can lead the way in providing a consistent model for defining fuzzy temporal inference rules for (some of) the commonly encountered imprecise temporal expressions (ITEs). To this end, we propose the fuzzification of the interval-based temporal logic in order to achieve a complete OWL-based reasoning for the fuzzy temporal built-ins. This is particularly important as it can enable the OWL/SWRL combination to enforce temporal semantics as well as handling vague temporal knowledge. Moreover, with the SWRL Query language (SQWRL) able to handle such temporal reasoning, querying temporal information from OWL ontologies will be highly improved.
\par 
For efficient representation and reasoning about the fuzzy temporal information encoded in FT-SWRL rules, we define the fuzzy times of the ITEs using carefully selected membership functions superimposed on their interval-based temporal definitions. The fuzzy membership functions were selected based on their correspondences to the imprecise temporal expressions using the weighted value (w) as the gauge of the temporal information as it approached the true value (T). However, we give some formularized restrictions to these weighted values within which the statements are found to be a close-enough representation of the temporal information. We focused on the frequently used ITEs found in the crops domain knowledge narratives --- as  earlier presented in Section~\ref{sec:motivation}. 

\subsection{SWRL-FT Built-ins Fuzzification}

Following the linguistic terminologies and variables definition in the SWRL-FTO ontology and Built-ins, we describe the \textit{Fuzzification} \cite{Miliauskaite2014, 4017811} of the FT-SWRL built-ins that will serve as translation rules during fuzzy temporal reasoning. In what follows, we chose suitable membership functions for some selected imprecise temporal expressions and demonstrate how we can generate their corresponding fuzzy values based on the membership functions.

\begin{table}[htbp]
	\centering
	\begin{tabular}{||p{4.3cm}|p{7.7cm}|}
		%{|c|c|c|}
		\hline 
		\textbf{Fuzzy temporal term} & \textbf{'about' (T)}, where: T = fuzzy duration interval. \\ 
		\hline  
		sameAs & Around (T), approximately (T), nearly (T).
		\\\hline 
		Super Class &	FuzzyModifier (annotation: ApproximationITEs)\\\hline
		Properties & hasWeightDegree, hasFuzzyTime, hasModifierFunction\\\hline
		Required arguments &	WeightDegree, FuzzyTime\\\hline
		Fuzzy MF &	Gaussian (Gaussmf)\\
		\hline
	\end{tabular} 
	\caption{Fuzzy temporal term \textit{about}}
	\label{tab:about}
\end{table}

\begin{figure}[tbph]
	\centering
	\includegraphics[width=0.85\linewidth, height=0.30\textheight]{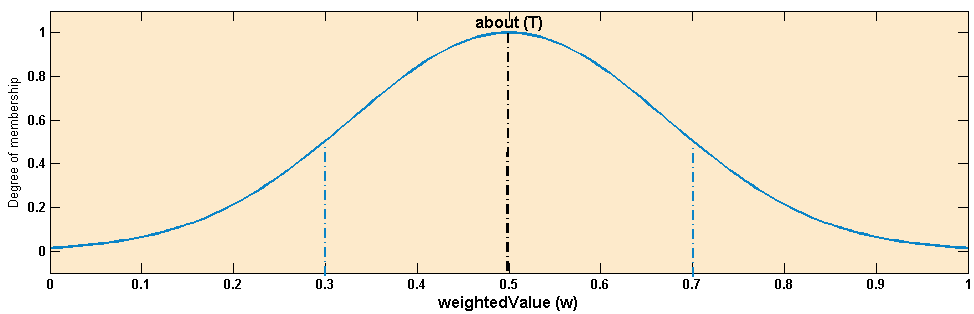}
	\caption{Membership Function for 'about (T)' Fuzzy Temporal Expression}
	\label{fig:aboutmf}
\end{figure}

As shown in Figure~\ref{fig:aboutmf}: Membership Function for 'about (T)' Fuzzy Temporal Expression, we use the Gaussian membership function (gaussmf) to define the fuzziness as a set over the about temporal expression as the universal set of discourse. Based on the semantic definition of the 'temporal approximation' keywords, such as `about', and `around', the use of such imprecise times is usually where the narrator refers to time units that are close to the exact time (of an event or process) and when such assumed times are up to a complete granularity. For example, the statement "Bambara beans germination time is \textit{around/about/approximately} 1 to 2 weeks". Here, the granularity (of weeks) is used to show that an event --- the germination of Bambara beans, may happen either in the first or the second week. This can basically be represented by the Gaussian MF (gaussmf), with the temporal value T = 7 days or 1st week as the peak-value (weight degree w = 1). The peak time can then be approached from either direction with increasing certainty (as 'w' tends to 1) until the actual time (truthTime T) of the event is reached. As such, the required information needed to model the approximation keyword will simply be 'the weighted degree of truth' of the information source.

The fuzzy time ($f_T$) for each ITE can then be calculated based on the assigned membership distribution function and weighted value. For the approximation ITEs, we calculate the minimum fuzzy time ($\min f_T$) and maximum fuzzy times ($\max f_T$) as the border-points for the resulting fuzzy temporal membership function as follows:

\begin{equation}\label{keya}
\min f_T = [T-(1-w)*T/2]
\end{equation}
\begin{equation}\label{key2}
\max f_T = [T+(1-w)*T/2]
\end{equation}
\begin{equation}\label{key3}
\min f_T < f_T < \max f_T 
\end{equation}

Where:
\begin{itemize}
	\item $T$ is the valid time unit in the temporal expression;
\item $w$ is the weighted truth degree of the expression (or information source);
\item $(1-w)*T/2$ is the distribution function for determining the fuzzy time based on the '$w$' values;
\item $f_T$ is the fuzzy time based on the weightedValue(w);
\item $\min f_T$  is the lower-bound fuzzy time for the \textit{about (T)} expression;
\item $\max f_T$  is the upper-bound fuzzy time for the \textit{about (T)} expression.
\end{itemize}

 Note that, the intuition in the distribution function is that the higher the degree of certainty (w), the closer the fuzzy times ($f_{T--})$ / $f_{T++})$) becomes to the actual valid time (T) on both sides.

With the above equations (\ref{keya} - \ref{key3}), we simplified the fuzzification of the ITE by calculating the possible minimum and maximum valid times for the expression. We use T/2 as a simplified distribution of the fuzzy variable (w) for the 'about (T)' expression, which implies that the fuzzy time can take values from $T - T/2$ for the possible times before T, to the $T + T/2$ possible timestamps after T. This is found to be consistent with our explanation that the about/around/approximately ITEs are commonly used to describe imprecise times (or events) that are within 1 or 2 granularities to the expected or precise time.
\par 
Example: Consider the expression; "Flowering time of Bambara nut is around 30 days from the date after sowing"
Therefore, the parameters are: T = 30 days, suggested values from the 'about MF' for w = (0.3 – 0.7).
Now assuming w = 0.4, then:
\begin{equation}\label{key4}
\min f_T = [30-(1-0.4)*30/2]=21\ days] \quad	\Rightarrow very\ early
\end{equation}  
and
\begin{equation}\label{key5}
\max f_T = [30+(1-0.4)*30/2]=39\ days] \quad	\Rightarrow very\ late
\end{equation}
 
The resulting parameters can be easily interpreted thus: \textit{"The flowering is early if it occurs before 30 days and after 21 days. It is late if it occurs after 30 days but before the 39th day} --- thereby enabling the assertions of fuzzy terms \textit{late} and \textit{early} in to the knowledge base. Moreover, as the keywords, 'before' and 'after' are already defined as part of Allen's interval algebra \cite{Allen1983}, therefore reasoning operations with other relevant data (e.g. other flowering times) can easily generate a consistent temporal network that can infer additional knowledge. Moreover, this approach, as explained earlier, can be easily applied on existing temporal ontologies by introducing the temporal fuzzification (through the ITE built-ins from FT-SWRL rules) to generate the available fuzzy times ($f_T$). Such modeling scenario help to confirm the  earlier assertion that FT-SWRL will not only allow managing fuzzy temporal information in OWL ontologies but also help to improve the utilization of existing temporal operators.
\par Using similar approach, we fuzzify other relevant ITEs such as 'few (T)', 'within (T)', 'before (T)', and 'after (T)' as shown below. These ITEs were selected as the first set of SWRL fuzzy temporal built-ins as they are the most frequent expressions (based on surveys presented in \cite{Rong2017}) found in domain knowledge descriptions and natural language processors. 

\begin{table}[htbp]
	\centering
	\begin{tabular}{||p{4.3cm}|p{7.7cm}|}
		%{|c|c|c|}
		\hline 
		\textbf{Fuzzy temporal term} & \textbf{'Within' (T)},  where: T = fuzzy duration interval or granularity (e.g. done within a week). \\ 
		\hline  
		sameAs & in less than (T), in under (T), at most (T), in no more than (T), etc.
		\\\hline 
		Super Class &	FuzzyModifier (annotation: Time closure operator)\\\hline
		Properties & hasWeightDegree, hasFuzzyTime, hasModifierFunction\\\hline
		Required arguments &	WeightDegree, FuzzyTime\\\hline
		Fuzzy MF &	Trapezoidal (trapmf)\\
		\hline
	\end{tabular} 
	\caption{Fuzzy temporal term \textit{within}}
	\label{tab:about}
\end{table}

\begin{figure}[btph]
	\centering
	\includegraphics[width=0.80\linewidth, height=0.25\textheight]{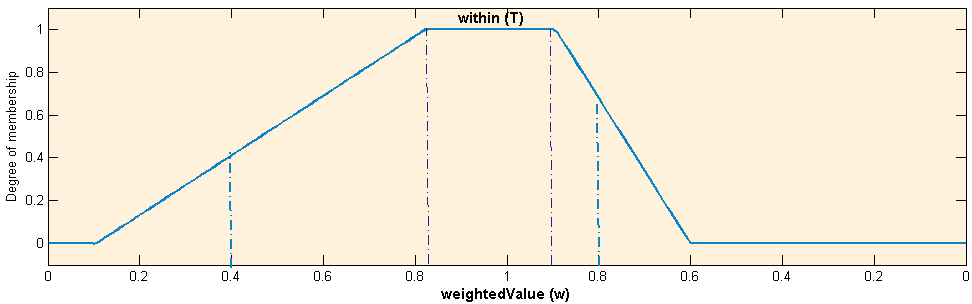}
	\caption[Membership Function for 'within (T)' Fuzzy Temporal Expression]{Membership Function for 'within (T)' Fuzzy Temporal Expression}
	\label{fig:withinmf}
\end{figure}

\paragraph{The 'within (T)' Built-in:}From the previous example, if the statement reads: "Bambara beans germinate \textit{within} 2 weeks from the date of sowing" it can be seen that the 'within' keyword is usually employed to express the maximum possible times that an event happens. Here we may use the '2 weeks' as the peak value. We use the Trapezoidal membership function (trapmf – see Fig. \ref{fig:withinmf}) to represent the progression of the fuzzy time as the weight-value (w) increases until the flat top --- where the valid-time (T) may be reached (i.e. $w = 1$). However, the sharp drop of the trapezoidal space function at the right-hand side corresponds to the small possible increment above the valid time ($T_++$). This follows the semantics of the 'within' operator where a small addition to a transaction time will still be valid e.g. '15 to 17 days' may still be referred as within 2 weeks in a fuzzified knowledge base (FKB). Note, however, a triangular membership function (trimf) can also be used, for simplicity, to represent the fuzzy space of the 'within' operation or where the valid time is a fuzzy instant time.

For the 'within' built-in, we calculate the minimum fuzzy time ($\min f_T$), the left-hand side of the trapezoid and the maximum fuzzy times ($\max f_T$) at the right, as the border points for the resulting fuzzy temporal value as follows:

\begin{equation}\label{key11}
\min f_T = [T-(1-w)*T]
\end{equation}  
\begin{equation}\label{key12}
\max f_T = [T+(1-w)*T]
\end{equation}
\begin{equation}\label{key13}
\min f_T < f_T < \max f_T 
\end{equation}

Where: $(1-w) * T$ 	is the distribution function for determining the fuzzy times based on the 'w' values.
Note: the intuition in the distribution function is that the higher the degree of certainty (w), the closer the fuzzy times ($f_{T--})$ / $f_{T++})$) becomes to the actual valid time (T) on both sides.

\begin{table}[hbtp]
	\centering
	\begin{tabular}{||p{4.3cm}|p{7.7cm}|}
		%{|c|c|c|}
		\hline 
		\textbf{Fuzzy temporal term} & \textbf{'Few (T)',},   where: T = granularity (e.g. completed in 'a few' days). \\ 
		\hline  
		sameAs & a few (T), a little (T), more or less (T), etc.
		\\\hline 
		Super Class &	FuzzyModifier \\\hline
		Properties & hasWeightDegree, hasFuzzyTime, hasModifierFunction\\\hline
		Required arguments & FuzzyTime, WeightDegree \\\hline
		Fuzzy MF &	Bell membership function (gbellmf)\\
		\hline
	\end{tabular} 
	\caption{Fuzzy temporal term \textit{few}}
	\label{tab:few}
\end{table}

\begin{figure}[tbph]
	\centering
	\includegraphics[width=0.7\linewidth, height=0.25\textheight]{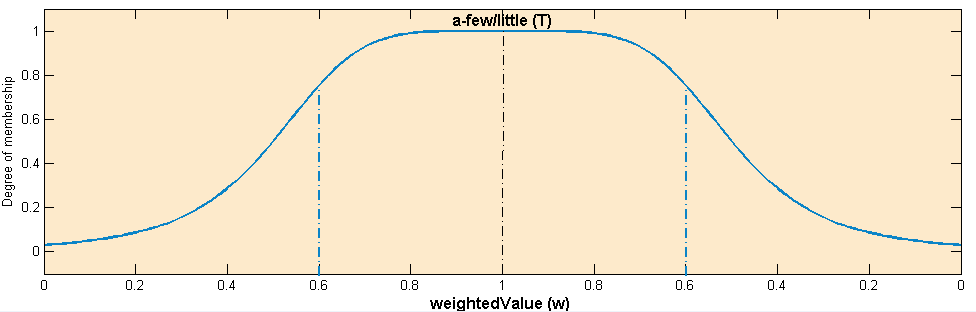}
	\caption[Membership Function for 'a-few (T)' Fuzzy Temporal Expression]{Membership Function for 'a-few (T)' Fuzzy Temporal Expression}
	\label{fig:fewmf}
\end{figure}

\paragraph{The 'few (T)' Built-in:}From the previous example, if the statement reads:"The Bambara beans will germinate in \textit{few weeks} if moderate rainfall continues", shows the use of the 'few' operator to express a flexible time increment without any specified amount or granularity. However, it usually represents small changes in time, which can sometimes be negligible. As discussed in \cite{Rong2017}, the use of this ITE is common in natural language narratives when the imprecise time referred to is very close to the actual time. We use the Gaussian bell membership function (gbellmf– see Fig. \ref{fig:fewmf}) --- which has gentle curves near the peak value, to represent the small progression of the fuzzy time as the weight value inchess towards the flat top (w = 1). 

For the \textit{few (T)} built-in, we calculate the minimum fuzzy time ($\min f_T$), the left-hand side of the bell and the maximum fuzzy times ($\max f_T$) at the right, as border points for the resulting fuzzy temporal value as follows:

\begin{equation}\label{key14}
\min f_T = [T-(1-w)*0.75 T]
\end{equation}  
\begin{equation}\label{key15}
\max f_T = [T+(1-w)*0.75 T ]
\end{equation}

Where: $(1-w) * 0.75 T$ the distribution function for determining the fuzzy times based on the ‘w’ values.
Similarly, as the degree of certainty (w) tends to 1, the fuzzy times ($f_{T--})$ / $f_{T++})$) tends to the actual valid time (T) from both sides.

\begin{table}[hbtp]
	\centering
	\begin{tabular}{||p{4.3cm}|p{7.7cm}|}
		%{|c|c|c|}
		\hline 
		\textbf{Fuzzy temporal term} & \textbf{'before (T)',},   where: T = date, time or granularity (e.g. happens before 3pm/7 days). \\ 
		\hline  
		sameAs & until, earlier than, previously, prior to, at most, etc.
		\\\hline 
		Super Class &	FuzzyModifier \\\hline
		Properties & hasWeightDegree, hasFuzzyTime, hasModifierFunction\\\hline
		Required arguments & FuzzyTime, WeightDegree (w = 0.6 – 1 )\\\hline
		Fuzzy MF &	S-membership function (smf)\\
		\hline
	\end{tabular} 
	\caption{Fuzzy temporal term \textit{before}}
	\label{tab:before}
\end{table}

\begin{figure}[tbph]
	\centering
	\includegraphics[width=0.6\linewidth, height=0.25\textheight]{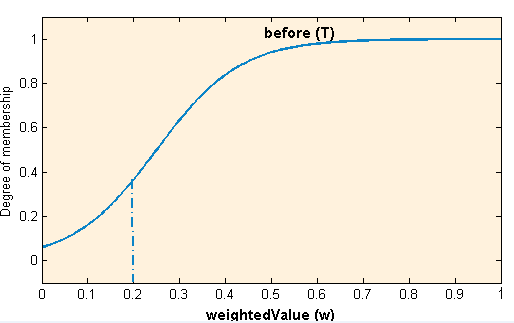}
	\caption[Membership Function for 'before (T)' Fuzzy Temporal Expression ]{Membership Function for 'before (T)' Fuzzy Temporal Expression }
	\label{fig:beforemf}
\end{figure}

\paragraph{The 'before (T)' Built-in:}
Consider the statement: "Bambara beans usually germinate \textit{before} 30 from the date of sowing”. This shows the use of 'before' ITE to express that germination do or will take place before 30 days (the peak period) after planting. However, it is not clear how close or far away the germination will be from the specified time. We use the S-membership function (smf --- see Fig. \ref{fig:beforemf}) to represent the progression of the fuzzy time with the increasing degree of truth from the bottom to the top of the S-function  --- where the fuzzy time may be equal to the valid time (T) i.e. w = 1. The continuous flat after the curve shows that we are not interested in times after the actual valid time ($T_{++}$). Hence the fuzziness is only on the left-hand side of the valid time to express (with some certainty) how close or far away the calculated fuzzy time is from the valid time. Which follows the semantics of the 'before' keyword --- used to safely express possible times prior to a known valid time. For example, 29, 20 or even 2 days may still be referred as 'before a month' in a fuzzy knowledge base based on the degree of certainty (w).

Similarly, we calculate the ($\min f_T$) and ($\max f_T$) as border points for the resulting fuzzy temporal value as follows:

\begin{equation}\label{key16}
\min f_T = [T-(1-w) * T/2]
\end{equation}  
\begin{equation}\label{key17}
\max f_T = T
\end{equation}

Note that as the S-function is designed here as the left-half of the Gaussian function, the minimum fuzzy times and calculated parameters are the same as the 'about' built-in. The exception being that 'before' built-in has a maximum fuzzy time of T as the degree of certainty (w) tends to 1. Hence, the consistency is preserved as $f_{T} \leqslant T$.

\begin{table}[htbp]
	\centering
	\begin{tabular}{||p{4.3cm}|p{7.7cm}|}
		%{|c|c|c|}
		\hline 
		\textbf{Fuzzy temporal term} & \textbf{'after (T)',},   where: T = date, time or granularity (e.g. arrive shortly 'after' 13:00 hours). \\ 
		\hline  
		sameAs &  Later than (T),  afterwards (T), subsequent to (T), etc.
		\\\hline 
		Super Class &	FuzzyModifier \\\hline
		Properties & hasWeightDegree, hasFuzzyTime, hasModifierFunction\\\hline
		Required arguments & FuzzyTime, WeightDegree (w = 0.6 – 1 )\\\hline
		Fuzzy MF &	Z-membership function (zmf)\\
		\hline
	\end{tabular} 
	\caption{Fuzzy temporal term \textit{after}}
	\label{tab:after}
\end{table}

\begin{figure}[tbph]
	\centering
	\includegraphics[width=0.7\linewidth, height=0.25\textheight]{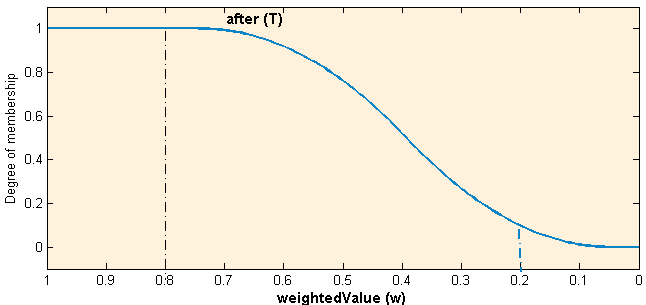}
	\caption[Membership Function for 'after (T)' Fuzzy Temporal Expression]{Membership Function for 'after (T)' Fuzzy Temporal Expression}
	\label{fig:aftermf}
\end{figure}

\paragraph{The 'after (T)' Built-in:} Consider the statement: "Bambara bean plant begins flowering \textit{after} 30 days from the date of sowing" shows the use of the 'after' as ITE to express that flowering happens after a month (the peak period). However, it cannot be ascertained how close or far away the flowering may start from the '30 days' after planting. In this case, we use the z-membership function (zmf – see Fig. \ref{fig:aftermf}) to represent the regression of the fuzzy time with the decreasing degree of truth from the top of the curve --- where the suggested valid time is known (w = 1). The continuous flat line before the curve shows that we are not interested in the time before the actual valid time. Hence the fuzziness is only on the right-hand side of the valid time ($T_{++} $). This also follows the natural semantics of the 'after' temporal expression, where it used to safely express possible transaction times that follows a known valid time. For example, 31, 32 or even 1000 days may still be referred to as 'after a month' in a fuzzy knowledge base. However, the built-in uses the weight value (w) as determining factor to safely express the possible times.

Hence, for the 'after' built-in operator, which is basically the opposite of 'before', we represent the ($\min f_T$) and ($\max f_T$) as border points for the resulting fuzzy temporal value as follows:

\begin{equation}\label{key18}
\min f_T = T
\end{equation}  
\begin{equation}\label{key19}
\max f_T = [T-(1-w) * T/2]
\end{equation}

Similarly, the maximum fuzzy time and other parameters are the same as the right-hand-side of the 'about' built-in with a mainimum fuzzy valid-time of $f_{T} \geqslant T$ as the degree of certainty (w) tends to 1. 

\section{Relevant Works}
\label{relevantworks}

We review relevant research approaches that handle representation and reasoning about fuzzy-temporal expressions, as well as language extensions that have been proposed to manage temporal expressions with uncertainties in the semantic web community.

\subsection{Fuzzy Temporal Representation and Reasoning (FTRR)}
\label{ftrr}
On the issue of modeling temporal uncertainties, there are two types of temporal uncertainties to be modeled: the imprecise dating of events and vague descriptions of time data \cite{Ziqiang227942}. As a domain modeling knowledge, the fuzzy temporal extension of SWRL aims to manage the latter by providing a formalism for representing imprecise time expressions.  Even though there is yet a formally accepted and standardized fuzzy-temporal reasoning system for the ontological knowledge bases, other logically validated fuzzy temporal reasoning systems have been proposed and developed over time. The following works on fuzzy temporal representation and reasoning offer an extensive literature and guide to developing new formalisms such as FT-SWRL.\par
A fuzzy temporal constraint satisfaction problem is defined in \cite{Deng:2009:FTL:1802007.1802063} as a new formalism for modeling flexibility and managing uncertainty into the interval-based temporal logic originally defined by J. Allen in [9]. The authors describe a reasoner based on Interval Constraint Network (ICN), which they claimed, can manage both the crisp and fuzzy temporal information containing uncertainties. Basic reasoning tasks described in this work involves the temporal consistency management and temporal query answering. The issue of consistency, as mentioned elsewhere, is important in all logical networks to achieve inference and this includes the fuzzy temporal networks. 
\newline

A similar approach was presented in \cite{lai2008fuzzy}, where the authors proposed a modeling and reasoning system for managing fuzzy temporal information commonly found in medical records. Here, an existing temporal reasoning system called 'TimeText', which allow representation of temporal information in clinical texts or narratives was extended to allow uncertain temporal data. The extension also proposed the use of fuzzy temporal constraint network (FTCN) with a proposed solution involving the three-state, staircase possibility distribution function. By exploring the complexity of possibility distributions in solving fuzzy temporal reasoning problems, the work relies heavily on the advances of \cite{Godo:1995} --- where the authors defined a propositional temporal language based on the fuzzy temporal constraints (FTCs). The proposal in \cite{Godo:1995} describes formal syntax and semantics based on possibilistic models and an inference mechanism based on FTC inference rules, with cited use cases in the medical domain. For more on FTRR systems, we refer interested readers to \cite{Schockaert4481150, Schockaert:2008, dubois2003fuzziness, 4017811} among others.

\subsection{Fuzzy Temporal Representation in Ontologies}
\label{FTRRinontologies}

Relevant works on temporal reasoning in the semantic web have focused more on the representation and reasoning of definite temporal information. Basic temporal ontologies are representations of time-stamps to allow modeling time-specific domain information. Most common temporal ontologies represent the basic metadata about time information; specifically, the 'points in time' data and others are simply domain-specific models for modeling crisp temporal data. Though, fewer efforts have aimed at providing consistent standards for reasoning on the temporal data. Notable efforts in this category includes: the SWRL Temporal \cite{Connor2011}, temporal OWL (tOWL) \cite{Milea:2012}, the OWL Time Ontology \cite{Hobbs2004} and its extensions such as in \cite{zhou2002reusable, Cox2016, pan2005temporal}. Others include: the Clinical Narratives Temporal Ontology (CNTRO) \cite{tao2010semantic} and CHRONOS \cite{chronos6984490} --- which, as the authors claimed, handles both qualitative as well as quantitate temporal facts. These works (and numerous others) have extensively discusses temporal modeling on the semantic web. However, real-life temporal information (including expert narratives) are usually inundated with non-crisp temporal description of events, procedures, etc. Nevertheless, the basic temporal models have provided groundworks for defining much of the fuzzy temporal representation models.
\newline
A notable fuzzy model for representing temporal uncertainties in ontologies is presented in \cite{Nagypal2003} --- which basically discusses the application of fuzzy temporal reasoning over historic data. Following a modular semantic approach, fuzzy set operations were employed to achieve the fuzzification of basic Allen intervals \cite{Allen1983} into fuzzy temporal intervals. However, due to its added flexibilities of second-order theory,  KAON2 \footnote{http://kaon2.semanticweb.org/} is selected as the target framework for the extension and not OWL directly. Moreover, to efficiently model temporal specifications using the model, an in-depth expertise in fuzzy interval logic is necessary. A factor we try to avoid by introducing the fuzzy temporal built-ins in our FT-SWRL specification. Thus giving users the simple natural language terminologies for modeling imprecise temporal information while hiding the technical implementation details. Moreover, importing the SWRL-FT ontology helps to achieve modeling consistency in the use of the fuzzy temporal SWRL rules. 
 
\section{Conclusions and Future Work}
\label{conclusion}

For the successful transformation of the current web of documents to the semantic web of interlinked data, there is the need for the domain modeling language of the semantic web to be able to represent all sorts of knowledge types. One of such important category is the uncertain-temporal data usually associated with human practices. With existing temporal ontology fragments, information can only be stored by associating each piece of knowledge with a fixed time stamp. However, when capturing domain expert's narratives involving imprecise temporal expressions, it becomes imperative to associate some sense of vagueness to the captured temporal facts for accurate representation of the domain knowledge. In this work, we presented a fuzzy temporal extension to the Semantic Web Rule Language (SWRL) called the Fuzzy Temporal SWRL (FT-SWRL), which combines fuzzy and temporal logics based on the valid-time temporal model \cite{Snodgrass1996adding}. The main contribution of the paper is twofold. The first contribution, described in section 3, is the introduction of a new set of ontological concepts and properties that allow to represent in RDF standard temporal propositions, which temporal specification is imprecise. The core set of these primitives are those used to express imprecise temporal expressions (ITEs). The second contribution, described in seciton 4, is constituted by a mathematical modelling, based on fuzzy set (fuzzy membership functions) for the ITEs. For each ITE, a specific membership function is proposed. In essence, the FT-SWRL provides a syntactical as well as semantic extension to the existing SWRL formalism for modeling imprecise temporal expressions (ITEs). The extension provides a consistent fuzzy-temporal model for managing imprecise temporal data in OWL ontologies using SWRL rules. It consists of two components: a fuzzy temporal SWRL ontology (SWRL-FTO) and a set of fuzzy temporal SWRL Built-ins. The ontology defines the language terminologies and variables to serves as reference standard for re-usability. While the built-ins allow for processing and reasoning about the imprecise temporal information modeled using the SWRL fuzzy temporal ontology. The paper also discussed the \textit{fuzzification} process (with examples) of some of the newly defined built-ins to support the semantic evaluation using carefully designed membership functions and inference rules. Example FT-SWRL rules were also presented to show the usability of the proposed fuzzy temporal extension model. 

In the future, we would like to incorporate more temporal vagueness into the FT-SWRL model and implement the complete reasoning system for the FT-SWRL rules. This can be achieved by modifying the existing SWRL-API Temporal model to support our newly-defined fuzzy-temporal operators. By leveraging the existing implementation mechanism for SWRL extensions, as described in \cite{OConnor2011}, the FT-SWRL model can be easily comprehended to allow modeling fuzzy temporal domain knowledge. A fuzzy temporal semantic web query language (FT-SQWRL) extension may also be considered.\par 
However, if a separate FT-SWRL reasoner is decided, reasoning can thus be achieved by linking the OWL ontologies holding the temporal propositions and their corresponding time-stamps, with the FT-SWRL model to generate  fuzzy temporal constraint network (FTCN). As discussed in \cite{Deng:2009:FTL:1802007.1802063} the fuzzified temporal expressions can then be solved as constraints satisfaction problems (CSPs) --- satisfied to a degree of truth based on their corresponding membership functions. 

\section*{References}



%\bibliographystyle{elsarticle-num} 
%\bibliography{ref2}

%\begin{thebibliography}{00}


%\bibitem{I. Horrocks, P. F. Patel-schneider, H. Boley, S. Tabet, B. Grosof, and M. Dean, “SWRL : A Semantic Web Rule Language Combining OWL and RuleML,” W3C Member submission 21. 2004.}

%\end{thebibliography}
\end{document}